\documentclass{article}
\usepackage{spconf,amsmath,graphicx}
\usepackage{booktabs}
\usepackage{listings}
\usepackage{xcolor}
\usepackage{amssymb}
\usepackage{float}
\usepackage{multirow}
\usepackage{siunitx}
\usepackage{setspace}

\usepackage{enumitem}
\setlist{nosep, leftmargin=14pt}

\usepackage{mwe} 

\newcommand{\eatme}[1]{ }

\title{A Reproducible Framework for Bias-Resistant Machine Learning on Small-Sample Neuroimaging Data}
\name{Jagan Mohan Reddy Dwarampudi$^{\dagger}$ \quad Jennifer L Purks$^{\star}$ \quad Joshua Wong$^{\star} $\quad Renjie Hu$^{\dagger\ddagger}$ \quad Tania Banerjee$^{\dagger\ddagger}$}
\address{Author Affiliation(s)}
\address{$^{\dagger}$ Department of Electrical and Computer Engineering, University of Houston \\
$^{\ddagger}$ Department of Information Science Technology, University of Houston \\
       $^{\star}$ Department of Neurology, College of Medicine, University of Florida}
\begin{document}
%
\maketitle

\begin{abstract}
We introduce a reproducible, bias-resistant machine learning framework that integrates domain-informed feature engineering, nested cross-validation, and calibrated decision-threshold optimization for small-sample neuroimaging data. Conventional cross-validation frameworks that reuse the same folds for both model selection and performance estimation yield optimistically biased results, limiting reproducibility and generalization. Demonstrated on a high-dimensional structural MRI dataset of deep brain stimulation cognitive outcomes, the framework achieved a nested-CV balanced accuracy of 0.660\,$\pm$\,0.068 using a compact, interpretable subset selected via importance-guided ranking. By combining interpretability and unbiased evaluation, this work provides a generalizable computational blueprint for reliable machine learning in data-limited biomedical domains.

\eatme{We introduce a reproducible, bias-resistant machine learning framework that integrates domain-informed feature engineering, nested cross-validation, and calibrated decision-threshold optimization for small-sample neuroimaging data. Conventional cross-validation frameworks that reuse the same folds for both model selection and performance estimation yield optimistically biased results, limiting reproducibility and generalization. Demonstrated on a high-dimensional structural MRI dataset of deep brain stimulation outcomes, the framework achieved a nested-CV balanced accuracy of \textbf{0.616~$\pm$~0.069 (check)} using only two interpretable features (check). By uniting interpretability, calibration, and unbiased evaluation, this work provides a generalizable computational blueprint for trustworthy machine learning in data-limited biomedical domains. }


\end{abstract}

\begin{keywords}
    Machine learning, Neuroimaging, Structural magnetic resonance imaging, Deep brain stimulation, Nested cross-validation
\end{keywords}

\section{Introduction}
\vspace{-2mm}
Machine learning applications in neuroimaging face the challenge of balancing model sophistication with methodological rigor. Datasets are typically small, feature spaces are high-dimensional, and subtle preprocessing or validation choices can unintentionally introduce bias~\cite{Varoquaux2017Assessing, Poldrack2020BestPractices, Rosenblatt2024DataLeakage}. Although MRI-derived features hold great promise for predicting clinical outcomes~\cite{artusi2020deep, lin2015biomarkers, hanganu2014mild}, limited cohort sizes and single-layer cross-validation schemes often lead to optimistically biased estimates that undermine reproducibility and generalizability~\cite{Varma2006, Cawley2010, Hastie2009Elements}. Furthermore, many pipelines rely on opaque, high-capacity models whose internal representations obscure anatomical meaning and hinder clinical translation. Conventional cross-validation, when reused for both model selection and performance estimation, can inflate error metrics by more than 20\%~\cite{Varma2006, Cawley2010}, highlighting the need for nested evaluation frameworks that yield unbiased, interpretable, and clinically meaningful results.

To address these challenges, we introduce a reproducible, bias-resistant, and modular machine-learning framework tailored for small-sample neuroimaging contexts. The framework unifies domain-aware feature engineering (FE), bias-controll-ed nested evaluation, and calibrated decision-threshold selection into a single coherent workflow. Rather than relying on end-to-end deep learning or generic dimensionality reduction, it begins with interpretable morphometric composites that capture neuroanatomical relationships (e.g., ventricle–brain ratio, gray/white-matter ratio, deep gray-matter volume). Model training and parameter optimization occur exclusively within nested cross-validation, while threshold search and probability calibration ensure consistent and unbiased decision boundaries across folds. Demonstrated on a structural MRI dataset predicting deep brain stimulation (DBS) outcomes~\cite{reich2022brain, mangone2020early, halpern2009cognition}, the framework exemplifies a principled and transparent approach that produces robust, deployment-ready classifiers with quantified generalization error and interpretable feature contributions

This paper makes three primary contributions:
\begin{enumerate}
    \item We introduce a rigorously leakage-controlled, fully reproducible machine-learning framework for small-sample neuroimaging, integrating strictly nested cross-validation, inner-loop probability calibration, and threshold optimization to ensure unbiased performance estimation.
    \item We develop a transparent, domain-informed feature-engineering pipeline that converts high-dimensional regional MRI labels into compact, biologically interpretable morphometric composites and asymmetries that preserve anatomical meaning.
    \item We provide a generalizable methodological template validated on DBS cognitive-outcome prediction, demonstrating how interpretable feature design and unbiased evaluation can be combined into a model-agnostic workflow suitable for any structured biomedical dataset with high feature dimensionality and limited sample size.
\end{enumerate}

\eatme{\item A reproducible, bias-controlled ML framework that integrates nested cross-validation with threshold optimization and probability calibration, yielding nearly unbiased performance estimates for small-sample neuroimaging data.
    \item A transparent feature-engineering pipeline that transforms high-dimensional regional MRI data into compact, biologically interpretable composites.
    \item A generalizable methodology demonstrated on DBS outcome prediction but applicable to any structured biomedical dataset characterized by limited sample size and high feature dimensionality.}
\vspace{-2mm}
\section{Methodology}\label{sec:method}
\vspace{-1mm}
This section describes the dataset, preprocessing procedures, and the key components of our analytical pipeline that: (i) builds domain-informed, interpretable features from structural MRI, and (ii) performs strictly nested cross-validation (CV) for unbiased model selection and performance estimation, and (iii) applies probability-threshold optimization to ensure stable operating points. All analyses used scikit-learn (v1.5)~\cite{pedregosa2011scikit} with fixed random seeds.

\vspace{1pt}
\noindent\textbf{Dataset and Preprocessing.}
We retrospectively analyzed structural MRI data from $n{=}332$ patients evaluated at the University of Florida Movement Disorders Center as part of their pre-surgical workup for deep brain stimulation (DBS). Each participant underwent neuropsychological testing and was rated using the UF Cognitive Risk Score (CRS)~\cite{kenney2020uf}, where a score of $3$ reflects average concern for post-surgical cognitive decline. Patients were dichotomized into ``low-risk'' ($\mathrm{CRS}<3$) and ``elevated-risk'' ($\mathrm{CRS}\ge3$) groups consistent with practice recommendations for DBS candidacy. T1-weighted scans were processed with \textit{FreeSurfer~\cite{dale1999cortical}} and \textit{SynthSeg~\cite{billot2023synthseg, desikan2006automated}} to obtain 109 regional volumes. Quality-control (QC) features generated by FreeSurfer were excluded from the analysis, and participants with failed or incomplete reconstructions (e.g., topological defects or missing segmentations) were removed prior to feature extraction. To enhance predictive signal, derived measures included volumetric fractions, ventricle--brain and gray/white-matter ratios, deep-gray-matter composites, bilateral asymmetries, and lobar aggregates. Global $z$-scaling was avoided to preserve biologically meaningful magnitudes. 

\vspace{1pt}
\noindent\textbf{Feature Engineering and selection.}
We derived interpretable composites to control for global and confounding effects and to encode known anatomical relationships: (i) regional volume fractions normalized by total intracranial volume (TIV), (ii) global atrophy surrogates (ventricle–brain and gray/white-matter ratios), (iii) deep gray-structure aggregates, and (iv) low-order interactions (e.g., age with ventricular fraction). Higher-level lobar summaries, asymmetry indices, and cortical descriptors were then formed.

\vspace{1pt}
\noindent\textbf{Model training and evaluation (baseline).}
All available features were retained without feature selection. 
For each dataset, we evaluated a panel of classifiers (Random Forest~\cite{breiman2001random}, Extra Trees~\cite{geurts2006extremely}, Gradient Boosting~\cite{friedman2001greedy}, k-Nearest Neighbors (KNN), Linear Discriminant Analysis (LDA), Multilayer Perceptron (MLP), Naive Bayes (NB), etc.) 
using a leakage-controlled nested CV protocol (outer $K{=}5$ folds; inner grid search within each outer-train split). 
In the inner loop, hyperparameters were tuned with stratified CV and refit using the average-precision criterion. 
The tuned model was then trained on the outer-train split, and a sigmoid (Platt~\cite{platt1999probabilistic}) calibrator was fit on held-out scores from the same training data. 
A single operating threshold $t^*$ was selected on calibrated outer-train predictions to maximize balanced accuracy (BA). 
Finally, the calibrated model and the train-selected threshold $t^*$ were applied to the untouched outer-test split for scoring. 
We report mean~$\pm$~SD across outer folds for BA (at $t^*$), AUC--ROC, AUPRC, Brier score, and expected calibration error (ECE)~\cite{niculescu2005predicting}.

\vspace{1pt}
\noindent\textbf{Leakage control.}
All transformations that can induce leakage (hyperparameter tuning, calibration, and threshold search) 
were confined strictly to the training side of each outer fold. 
Outer-test data were never used for model selection. 
No correlation filtering, importance ranking, or wrapper selection was performed; 
tree-based models were trained on the full feature set without input scaling.

\vspace{1pt}
\noindent\textbf{Nested evaluation and thresholding.}
Model evaluation followed the nested (CV) design described above, 
consisting of a 5-fold stratified outer CV for unbiased performance estimation 
and a 3-fold inner CV for hyperparameter optimization. 
All classifiers were trained and tuned within this framework (see Table~\ref{tab:synthseg-ba}), 
with class-weighted Random Forests achieving the most consistent performance across folds. 
For each outer fold, a sigmoid (Platt) calibrator was trained on the outer-train scores, 
and an optimal decision threshold $t^*$ was selected on the calibrated probabilities 
to maximize balanced accuracy (BA). 
The resulting threshold was then applied to the corresponding outer-test predictions. 
All calibration and threshold searches were confined to the inner training data to prevent leakage, 
ensuring that the reported metrics (BA, AUC--ROC, AUPRC, Brier, and ECE) reflect true out-of-sample generalization.

\vspace{1pt}
\noindent\textbf{Metric and Reproducibility.}
Performance was summarized solely by balanced accuracy (BA),
computed on each outer test fold and reported as mean~$\pm$~standard deviation. To evaluate the effect of probability calibration, area under the receiver operating characteristic curve (AUC–ROC) was additionally computed on out-of-fold probabilities for both uncalibrated and calibrated models. The entire pipeline (preprocessing, feature engineering, selection, thresholding, and evaluation) is scripted with fixed versions and \mbox{random\_state\,=\,42} to enable exact reruns from a single configuration file. Code will be released upon publication.

\begin{table}[t]
    \footnotesize
    \centering
    \renewcommand{\arraystretch}{1.2}
    \resizebox{1.0\columnwidth}{!}{%
        \begin{tabular}{l cc cc}
            \toprule
            \multirow{2}{*}{\textbf{Model}} &
            \multicolumn{2}{c}{\textbf{SynthSeg original}} &
            \multicolumn{2}{c}{\textbf{SynthSeg FE}} \\
            \cmidrule(lr){2-3}\cmidrule(lr){4-5}
            & \textit{BA (mean $\pm$ SD)} & \textit{Med.\ thr.} & \textit{BA (mean $\pm$ SD)} & \textit{Med.\ thr.} \\
            \midrule
            AdaBoost            & 0.670 $\pm$ 0.047 & 0.46 & 0.679 $\pm$ 0.049 & 0.47 \\
            Bagging             & 0.685 $\pm$ 0.050 & 0.38 & 0.681 $\pm$ 0.052 & 0.38 \\
            Bernoulli NB        & 0.628 $\pm$ 0.046 & 0.28 & 0.676 $\pm$ 0.046 & 0.35 \\
            Decision Tree       & 0.620 $\pm$ 0.053 & 0.27 & 0.614 $\pm$ 0.052 & 0.33 \\
            Extra Trees         & 0.696 $\pm$ 0.042 & 0.39 & \textbf{0.713 $\pm$ 0.050} & 0.39 \\
            Gaussian NB         & 0.649 $\pm$ 0.046 & 0.50 & 0.671 $\pm$ 0.046 & 0.65 \\
            Gradient Boosting   & \underline{0.700 $\pm$ 0.050} & 0.28 & 0.680 $\pm$ 0.046 & 0.28 \\
            KNN                 & 0.653 $\pm$ 0.047 & 0.32 & 0.664 $\pm$ 0.043 & 0.31 \\
            LDA                 & 0.638 $\pm$ 0.044 & 0.39 & 0.691 $\pm$ 0.040 & 0.40 \\
            Logistic Regression & 0.635 $\pm$ 0.041 & 0.37 & 0.697 $\pm$ 0.041 & 0.39 \\
            MLP Classifier      & 0.627 $\pm$ 0.052 & 0.42 & 0.673 $\pm$ 0.054 & 0.42 \\
            Multinomial NB      & 0.620 $\pm$ 0.039 & 0.39 & 0.638 $\pm$ 0.041 & 0.39 \\
            Random Forest       & \textbf{0.703 $\pm$ 0.047} & 0.41 & \underline{0.705 $\pm$ 0.049} & 0.39 \\
            \bottomrule
        \end{tabular}
    }
    \caption{Balanced test accuracy (BA; mean $\pm$ SD across outer folds) and median decision threshold (\emph{Med.\ thr.}) for SynthSeg original vs.\ feature-engineered (FE) feature sets. Highest BA per column in \textbf{bold}; second-highest \underline{underlined}. Thresholds rounded to two decimals.}
    \label{tab:synthseg-ba}
    \vspace{-5mm}
\end{table}

\begin{table}[t]
    \footnotesize
    \centering
    \renewcommand{\arraystretch}{1.2}
    \resizebox{1.0\columnwidth}{!}{%
        \begin{tabular}{l cc cc}
            \toprule
            \multirow{2}{*}{\textbf{Model}} &
            \multicolumn{2}{c}{\textbf{SynthSeg original (calibrated)}} &
            \multicolumn{2}{c}{\textbf{SynthSeg FE (calibrated)}} \\
            \cmidrule(lr){2-3}\cmidrule(lr){4-5}
            & \textit{BA (mean $\pm$ SD)} & \textit{Med.\ thr.} & \textit{BA (mean $\pm$ SD)} & \textit{Med.\ thr.} \\
            \midrule
            AdaBoost            & 0.601~$\pm$~0.059 & 0.39 & 0.607~$\pm$~0.056 & 0.39 \\
            Bagging             & 0.617~$\pm$~0.064 & 0.39 & 0.625~$\pm$~0.049 & 0.40 \\
            Bernoulli NB        & 0.583~$\pm$~0.050 & 0.38 & 0.598~$\pm$~0.054 & 0.38 \\
            Decision Tree       & 0.586~$\pm$~0.055 & 0.35 & 0.614~$\pm$~0.066 & 0.34 \\
            Extra Trees         & 0.628~$\pm$~0.062 & 0.39 & 0.634~$\pm$~0.058 & 0.39 \\
            Gaussian NB         & 0.589~$\pm$~0.058 & 0.39 & 0.584~$\pm$~0.057 & 0.40 \\
            Gradient Boosting   & 0.622~$\pm$~0.054 & 0.39 & 0.643~$\pm$~0.050 & 0.38 \\
            KNN                 & 0.549~$\pm$~0.040 & 0.38 & 0.578~$\pm$~0.061 & 0.37 \\
            LDA                 & 0.556~$\pm$~0.058 & 0.39 & 0.619~$\pm$~0.049 & 0.39 \\
            Logistic Regression & 0.574~$\pm$~0.070 & 0.40 & 0.631~$\pm$~0.051 & 0.39 \\
            MLP Classifier      & 0.552~$\pm$~0.058 & 0.39 & 0.569~$\pm$~0.063 & 0.39 \\
            Multinomial NB      & 0.517~$\pm$~0.045 & 0.10 & 0.544~$\pm$~0.056 & 0.39 \\
            Random Forest       & \textbf{0.656~$\pm$~0.056} & 0.40 & \textbf{0.660~$\pm$~0.068} & 0.39 \\
            \bottomrule
        \end{tabular}
    }
    \caption{Balanced test accuracy (BA; mean~$\pm$~SD across outer folds) and median calibrated decision threshold (\emph{Med.\ thr.}) for SynthSeg original vs.\ feature-engineered (FE) feature sets. Highest BA per column in \textbf{bold}; second-highest \underline{underlined}. Thresholds rounded to three decimals.}
    \label{tab:synthseg-ba-calibrated}
\end{table}
\begin{table*}[t]
    \small
    \centering
    \renewcommand{\arraystretch}{1.2}
    \resizebox{0.8\textwidth}{!}{%
        \begin{tabular}{l cc cc}
            \toprule
            \multirow{2}{*}{\textbf{Evaluation strategy}} &
            \multicolumn{2}{c}{\textbf{Random Forest (BA)}} &
            \multicolumn{2}{c}{\textbf{Extra Trees (BA)}} \\
            \cmidrule(lr){2-3} \cmidrule(lr){4-5}
            & \textit{Original} & \textit{FE} & \textit{Original} & \textit{FE} \\
            \midrule
            Train/Test split
            & 0.574~$\pm$~0.057 & 0.578~$\pm$~0.050
            & 0.576~$\pm$~0.057 & 0.567~$\pm$~0.050 \\

            Train/Test split + Grid Search CV
            & 0.587~$\pm$~0.056 & 0.581~$\pm$~0.047
            & 0.578~$\pm$~0.045 & 0.575~$\pm$~0.047 \\

            Naive CV
            & 0.582~$\pm$~0.019 & 0.577~$\pm$~0.017
            & \underline{0.586~$\pm$~0.015} & \underline{0.580~$\pm$~0.022} \\

            Naive CV + Grid Search CV
            & \underline{0.592~$\pm$~0.027} & \underline{0.582~$\pm$~0.015}
            & 0.578~$\pm$~0.020 & 0.574~$\pm$~0.021 \\

            \textbf{Proposed method (nested CV + Platt scaling)}
            & \textbf{0.656~$\pm$~0.056} & \textbf{0.660~$\pm$~0.068}
            & \textbf{0.628~$\pm$~0.062} & \textbf{0.634~$\pm$~0.058} \\
            \bottomrule
        \end{tabular}
    }
    \caption{Balanced accuracy (BA; mean~$\pm$~SD across outer folds) for the two top-performing models under different evaluation strategies. Columns show Original vs.\ feature-engineered (FE) feature sets within each model. The proposed nested CV incorporates sigmoid probability calibration (Platt scaling). \textbf{Bold}~= highest per column; \underline{underline}~= second-highest.}
    \label{tab:rf-et-matrix}
    \vspace{-5mm}
\end{table*}

\begin{table}[t]
    \scriptsize
    \centering
    \renewcommand{\arraystretch}{1.1}
    \setlength{\tabcolsep}{4pt}
    \resizebox{1.0\columnwidth}{!}{%
        \begin{tabular}{l c l c}
            \toprule
            \multicolumn{2}{c}{\textbf{SynthSeg original (top features)}} &
            \multicolumn{2}{c}{\textbf{SynthSeg FE (top features)}} \\
            \cmidrule(lr){1-2}\cmidrule(lr){3-4}
            \textbf{Feature} & \textbf{Imp.} &
            \textbf{Feature} & \textbf{Imp.} \\
            \midrule
            ctx-rh-paracentral\_fracs & 0.080 & ctx-rh-paracentral & 0.100 \\
            ctx-lh-superiorfrontal\_fracs & 0.072 & right pallidum & 0.090 \\
            right pallidum & 0.065 & left inferior lateral ventricle & 0.087 \\
            left pallidum & 0.058 & right inferior lateral ventricle & 0.087 \\
            right inferior lateral ventricle\_fracs & 0.058 & left lateral ventricle & 0.081 \\
            ctx-rh-bankssts\_fracs & 0.055 & ctx-rh-caudalmiddlefrontal & 0.079 \\
            4th ventricle\_fracs & 0.053 & right lateral ventricle & 0.077 \\
            ctx-rh-medialorbitofron\_asym & 0.051 & brain-stem & 0.074 \\
            left inferior lateral ventricle\_fracs & 0.050 & ctx-rh-lingual & 0.073 \\
            left cerebellum white matter\_fracs & 0.049 & ctx-lh-supramarginal & 0.072 \\
            ctx-lh-paracentral\_fracs & 0.049 & ctx-lh-lateraloccipital & 0.071 \\
            left lateral ventricle\_fracs & 0.047 & ctx-lh-precentral & 0.067 \\
            ctx-lh-medialorbitofron\_asym & 0.046 & ctx-rh-frontalpole & 0.062 \\
            right cerebellum white matter\_fracs & 0.046 & left accumbens area & 0.058 \\
            ctx-lh-posteriorcingulate\_fracs & 0.043 & ctx-lh-isthmuscingulate & 0.057 \\
            ctx-lh-middletempo\_asym & 0.043 & left caudate & 0.055 \\
            \bottomrule
        \end{tabular}
    }
    \caption{Top feature subsets and corresponding Random Forest importances averaged across folds. Left: original SynthSeg regions; Right: feature-engineered (FE) composites/asymmetries. Importances are mean values (rounded to three decimals).}
    \label{tab:selected-feature-importances}
    \vspace{-5mm}
\end{table}

\vspace{-2mm}
\section{Results}\label{sec:results}
\vspace{-1mm}
Model performance was evaluated under nested CV. 
As summarized in Tables~\ref{tab:synthseg-ba} and~\ref{tab:synthseg-ba-calibrated}, domain-informed feature engineering produced small but consistent gains across most classifiers, with \emph{Extra Trees} and \emph{Random Forest} models achieving the highest BA on the feature-engineered (FE) representation. 
After Platt scaling, model ranking remained consistent and probabilities became better calibrated, with median decision thresholds converging near~0.39 across folds. 

\vspace{-2mm}
\subsection{Nested cross-validated performance}
\vspace{-1mm}

Within the leakage-controlled nested CV framework, consistent and unbiased generalization performance was observed across classifiers. After applying Platt scaling, the \emph{Random Forest} achieved the highest BA on both datasets, (Table~\ref{tab:synthseg-ba-calibrated}). The \emph{Extra Trees} classifier performed comparably.  

Feature engineering yielded a modest but consistent improvement of approximately +0.01 in balanced accuracy across models, demonstrating that the aggregated anatomical composites enhanced discriminative signal without increasing model complexity. In contrast, naïve CV and single train--test evaluations in earlier baselines produced lower and more variable accuracies (approximately 0.57--0.59~BA), reaffirming that the nested design mitigates optimistic bias introduced when hyperparameter tuning and performance estimation occur on overlapping data folds.

\vspace{-2mm}
\subsection{Feature engineering and threshold stability}
\vspace{-1mm}
Across both classifiers, the feature-engineered (FE) representation consistently matched or slightly exceeded the performance of the original regional volumes (Table~\ref{tab:synthseg-ba-calibrated}). After Platt scaling, BA improved modestly by approximately~+0.004 for the \emph{Random Forest} and by~+0.006 for the \emph{Extra Trees} model. These small but consistent gains indicate that compact, domain-informed morphometric composites preserve discriminative signal while reducing redundancy in the high-dimensional feature space, enabling more stable generalization without increasing model complexity.

Within each outer fold, probability thresholds were optimized on the inner validation (calibration) set to maximize BA using the sigmoid-calibrated probabilities. The resulting thresholds were tightly concentrated around a median value of $t = 0.39$ (interquartile range~$\approx 0.01$), yielding consistent fold-level performance of 0.66~$\pm$~0.01~BA for the FE representation (Table~\ref{tab:foldwise-thr-ba-fe}). When applied to the original SynthSeg features, fold-level accuracies remained stable at 0.63~$\pm$~0.05~BA with nearly identical thresholds ($t = 0.40$~$\pm$~0.00; Table~\ref{tab:synthseg-ba-calibrated}). Using the median threshold as a fixed decision boundary across outer folds preserved balanced accuracy and minimized inter-fold variability, confirming the robustness of the calibrated operating point.

\begin{figure}[t]
    \centering
    \includegraphics[width=\columnwidth]{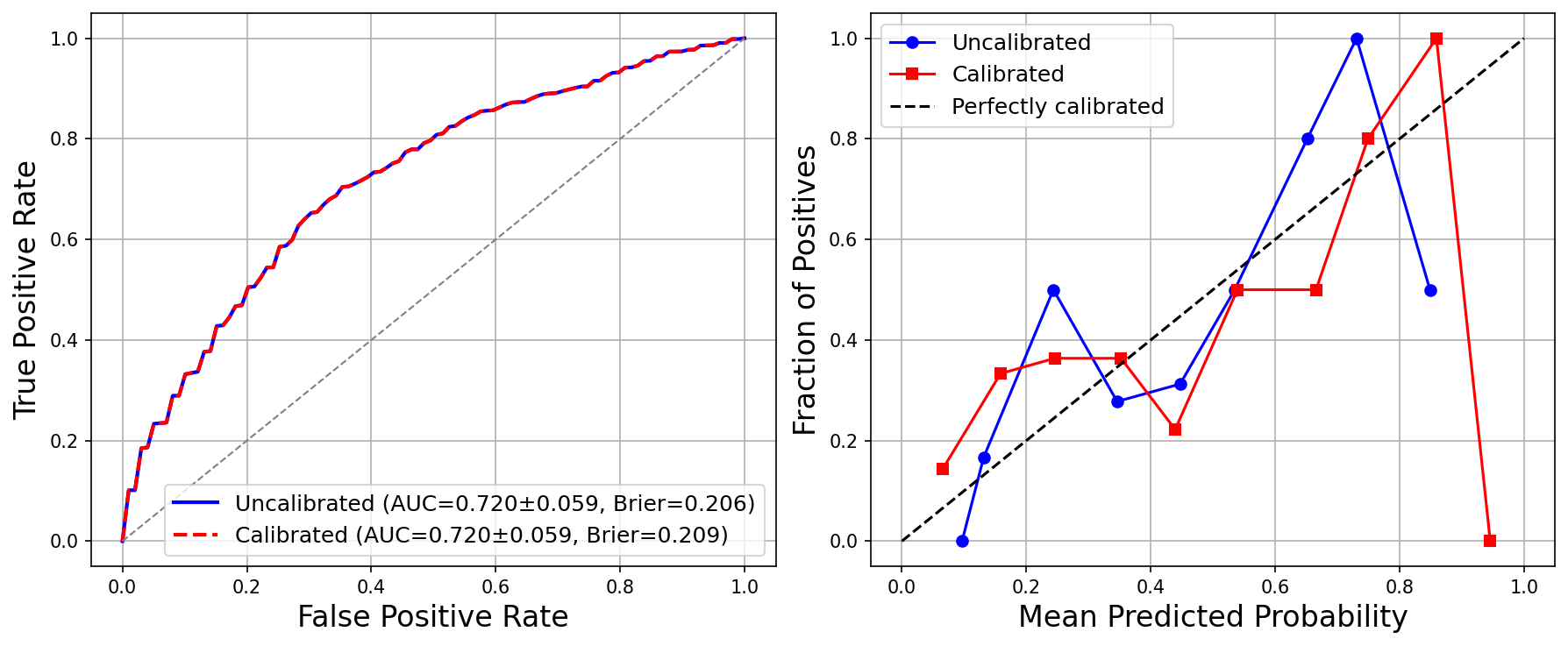}
    \caption{Receiver-operating characteristic (ROC) curve and calibration reliability diagram for the Random Forest trained on feature-engineered SynthSeg features under nested CV with sigmoid calibration (Platt scaling). Mean AUC--ROC~$\approx$~0.72, Brier~$\approx$~0.22, and ECE~$\approx$~0.13. The mean AUC–ROC $\approx$ 0.72 reflects ranking performance across thresholds, whereas the balanced accuracy in Table~\ref{tab:synthseg-ba-calibrated} reports accuracy at the optimized decision threshold.}
    \label{fig:roc-calibration}
\end{figure}

\vspace{-2mm}
\subsection{Calibration and bias control}
\vspace{-1mm}

Figure~\ref{fig:roc-calibration} illustrates model discrimination and probability calibration under the nested CV protocol with Platt scaling. 
Although the reliability curve deviates slightly from the ideal diagonal, the expected calibration error (ECE~$\approx$~0.13--0.14) and Brier score ($\approx$~0.22) indicate that the probability estimates are well calibrated. 
These minor departures likely reflect limited fold size rather than systematic bias or overconfidence. 
Calibration improved probabilistic consistency without altering overall discrimination (AUC--ROC~$\approx$~0.70--0.72); Precision--Recall behavior followed a similar pattern (AUPRC~$\approx$~0.59--0.62$)$ and is therefore omitted for brevity.

Although both AUC--ROC and balanced accuracy (BA) quantify classifier performance, they measure distinct properties. 
AUC--ROC ($\approx$\,0.70--0.72) reflects ranking ability across all thresholds, while BA ($\approx$\,0.65--0.66) captures accuracy at the single, optimized operating threshold ($t \approx 0.39$) chosen for deployment. 
Hence, the slightly higher AUC values reflect integration across thresholds rather than inflated performance, confirming consistent ranking and threshold-specific accuracy.

Relative to single-layer or naïve cross-validation (CV), where the same folds are reused for both hyperparameter tuning and performance estimation, the proposed nested protocol substantially reduced estimation bias and improved reproducibility. 
All feature selection, probability calibration, and threshold optimization were confined strictly to the inner loop, while the outer loop provided an unbiased estimate of generalization performance. 
Under this rigorous evaluation, the \emph{Random Forest} achieved balanced accuracies of 0.656~$\pm$~0.056 on the original SynthSeg features and 0.660~$\pm$~0.068 on the feature-engineered (FE) representation, compared with the 0.57--0.59~BA range obtained under naïve or single train--test evaluation (Table~\ref{tab:rf-et-matrix}). 
Fold-level accuracies exhibited low variability (standard deviation~$\approx$~0.05) and stable decision thresholds (median~$t = 0.39 \pm 0.01$), confirming the robustness of the calibrated operating point. 
By separating model selection and evaluation, the nested design eliminated optimistic bias and produced reproducible, trustworthy estimates of out-of-sample performance.

\begin{table}[t]
    \small
    \centering
    \renewcommand{\arraystretch}{1.2}
    \setlength{\tabcolsep}{6pt}
    \begin{tabular}{ccc}
        \toprule
        \textbf{Fold} & \textbf{Optimal Threshold} & \textbf{Fold BA} \\
        \midrule
        1 & 0.40 & 0.67 \\
        2 & 0.40 & 0.66 \\
        3 & 0.39 & 0.66 \\
        4 & 0.39 & 0.68 \\
        5 & 0.39 & 0.62 \\
        \midrule
        \textbf{Median $\pm$ IQR} & \textbf{0.39 $\pm$ 0.01} & \textbf{0.66 $\pm$ 0.01} \\
        \bottomrule
    \end{tabular}
    \caption{Optimal decision thresholds and fold-level balanced accuracy (BA) for the \textbf{proposed method} using the \textbf{feature-engineered SynthSeg} feature set. The summary row reports the median~$\pm$~interquartile range (IQR).}
    \label{tab:foldwise-thr-ba-fe}
    \vspace{-5mm}
\end{table}

\vspace{-2mm}
\section{Conclusion}
\vspace{-1mm}


We presented a reproducible framework for bias-resistant learning in small-sample neuroimaging that integrates domain-informed feature engineering with strictly nested cross-validation, probability calibration, and threshold optimization. Applied to structural MRI for deep brain stimulation (DBS) cognitive outcome prediction, the approach achieved a nested-CV balanced accuracy of approximately \(0.66 \pm 0.06\) and an AUC--ROC of \(0.72 \pm 0.06\), with stable operating thresholds (median \(t \approx 0.39\)). In contrast, naïve evaluation strategies yielded \(0.57\)–\(0.59\) balanced accuracy, highlighting the magnitude of optimism bias without leakage control.

The study was limited by its single-site design and moderate sample size, reflecting clinic-specific labeling practices and motivating conservative interpretation of results. Nevertheless, the framework is model- and modality-agnostic, preserves interpretability, and yields calibrated, deployment-ready decision rules. These properties support transparent benchmarking and provide a foundation for future multi-site validation and integration into clinical decision-support workflows for DBS evaluation and other high-dimensional, low-\(n\) biomedical applications.

\vspace{-4mm}
\section{Compliance with Ethical Standards}
\vspace{-2mm}
This retrospective study used de-identified structural MRI and clinical labels from patients evaluated for deep brain stimulation at the University of Florida Movement Disorders Center. The protocol was approved by the University of Florida Institutional Review Board (IRB) in accordance with the Declaration of Helsinki. Given the retrospective use of de-identified data, the requirement for informed consent was waived as per IRB determination. 

\vspace{-2mm}
\section{Conflicts of Interest}
\vspace{-2mm}
This work was supported in part by the Center for Transformative Pathology and Health (CTPH) under UM1TR004539. The authors have no relevant financial or non-financial interests to disclose.

\vspace{-2mm}
\def\IEEEbibitemsep{0pt plus .5pt}
\begin{spacing}{0.5}
\bibliographystyle{IEEEbib}
\bibliography{refs}
\end{spacing}

\end{document}